\documentclass[conference]{IEEEtran}
\IEEEoverridecommandlockouts    
\usepackage[margin=0.75in, top=1in]{geometry}
\usepackage{textcomp}
\def\BibTeX{{\rm B\kern-.05em{\sc i\kern-.025em b}\kern-.08em
    T\kern-.1667em\lower.7ex\hbox{E}\kern-.125emX}}
\usepackage{multirow}
\usepackage{lipsum}
\usepackage{amsmath}
\usepackage{amssymb}
\usepackage[ruled,vlined]{algorithm2e}
\usepackage{graphicx}
\usepackage{amsfonts}%
\usepackage{gensymb}%
\usepackage{mathrsfs}%
\usepackage{xcolor}%
\usepackage{textcomp}%
\usepackage{manyfoot}%
\usepackage{booktabs}%
\usepackage{placeins}
\usepackage{siunitx}
\usepackage{flushend}
\usepackage{hyperref}
\graphicspath{{./Figures/}}

\title{\LARGE \bf General-purpose LLMs as Models of Human Driver Behavior: \\ The Case of Simplified Merging}

\author{
	\parbox{\textwidth}{%
		\centering
		Samir H.A. Mohammad$^{1}$, Wouter Mooi$^{2}$, Arkady Zgonnikov$^{2}$%
	}%
    \thanks{$^{1}$Department of Transport and Planning, Delft University of Technology, $^2$Department of Cognitive Robotics, 2628CD Delft, The Netherlands. Corresponding author: {\tt\small s.h.a.mohammad@tudelft.nl}}%
}

\begin{document}
	
	\maketitle
    \footnotetext[1]{\textbf{Supplementary Information, Data and Software Availability}
    The software and simulated data as well as supplementary information supporting this paper are available online:
    \url{https://github.com/shamohammad/General-purpose-LLM-Merging}.}
	\thispagestyle{empty}
	\pagestyle{empty}
	
\begin{abstract}
Human behavior models are essential as behavior references and for simulating human agents in virtual safety assessment of automated vehicles (AVs), yet current models face a trade-off between interpretability and flexibility. General-purpose large language models (LLMs) offer a promising alternative: a single model potentially deployable without parameter fitting across diverse scenarios. However, what LLMs can and cannot capture about human driving behavior remains poorly understood. We address this gap by embedding two general-purpose LLMs (OpenAI o3 and Google Gemini 2.5 Pro) as standalone, closed-loop driver agents in a simplified one-dimensional merging scenario and comparing their behavior against human data using quantitative and qualitative analyses. Both models reproduce human-like intermittent operational control and tactical dependencies on spatial cues. However, neither consistently captures the human response to dynamic velocity cues, and safety performance diverges sharply between models. A systematic prompt ablation study reveals that prompt components act as model-specific inductive biases that do not transfer across LLMs. These findings suggest that general-purpose LLMs could potentially serve as standalone, ready-to-use human behavior models in AV evaluation pipelines, but future research is needed to better understand their failure modes and ensure their validity as models of human driving behavior.
\end{abstract}
	
\section{Introduction} \label{sec:introduction}
Automated vehicles (AVs) are expected to yield substantial societal benefits, including improved safety, enhanced mobility, and positive economic effects \cite{norman-lopez_economy-wide_2026}. Realizing these promises requires rigorous safety assessment, both on-road and virtual, to inform deployment and certification decisions. The rigor of virtual AV safety assessment depends strongly on human behavior models in at least two ways. First, human driver behavior models serve as \textit{behavior references}, providing benchmarks against which AV behavior can be compared~\cite{bargman_human_2025}. Second, they populate \textit{traffic simulation environments} to evaluate AV performance in encounters with human road users \cite{wang_application_2024}. In both cases, accurate human behavior models are essential for the validity of AV safety assessment.

Models of human driving behavior used for AV safety assessment can broadly be divided into mechanistic and data-driven approaches (although recently hybrid models are gaining prominence~\cite{schumann_using_2023, wang_modeling_2025}). Mechanistic models aim to capture the principles underlying driving behavior and offer interpretable, fine-grained explanations of human driving~\cite{markkula_explaining_2023, siebinga2024model, schumann_active_2025}. However, they often lack flexibility and require substantial adaptation for individual driving scenarios. Data-driven models, in contrast, capture high-dimensional behavioral patterns across diverse scenarios through learning from large-scale trajectory data (e.g.,~\cite{cornelisse_building_2025, guo_decompgail_2026}). Especially with the emergence of large language models (LLMs) and related generative AI techniques, data-driven models have gained unprecedented flexibility, enabling the generation of plausible, human-like driving behavior across varied scenarios. For example, next-token prediction approaches inspired by LLM principles represent driving as a sequence of motion tokens and use generative models, often diffusion-based, trained on tokenized trajectories to synthesize behavior (e.g., \cite{philion_trajeglish_2024, zhang_trajtok_nodate}).

Besides inspiring architectures of driver models, general-purpose LLMs are also increasingly incorporated directly into driving behavior models. For instance, a growing body of work has explored the use of pre-trained, general-purpose LLMs as modules for open-loop trajectory prediction architectures~\cite{xu_trajectory_2025}. However, models trained in an open-loop fashion can diverge from realistic behavior when deployed in closed-loop simulations~\cite{montali_waymo_2023}, which are essential for AV evaluation. In closed-loop traffic simulation, LLMs have also been used in various roles. SceneDiffuser++ \cite{tan_scenediffuser_2025} employs an LLM for constraint generation within a generative world model, while language-guided diffusion approaches use LLMs as prompting interfaces for motion generators \cite{zhong_language-guided_2023, tan_promptable_2024}. Motion-LLaVA \cite{li_womd-reasoning_2025} integrates an LLM backbone for reasoning over encoded motion representations. These works demonstrate the versatility of LLMs, but often embed them within complex pipelines, making it difficult to isolate their standalone capabilities as behavior models. 

\begin{figure*}[!htbp]
    \centering
    \includegraphics[width=0.7\textwidth]{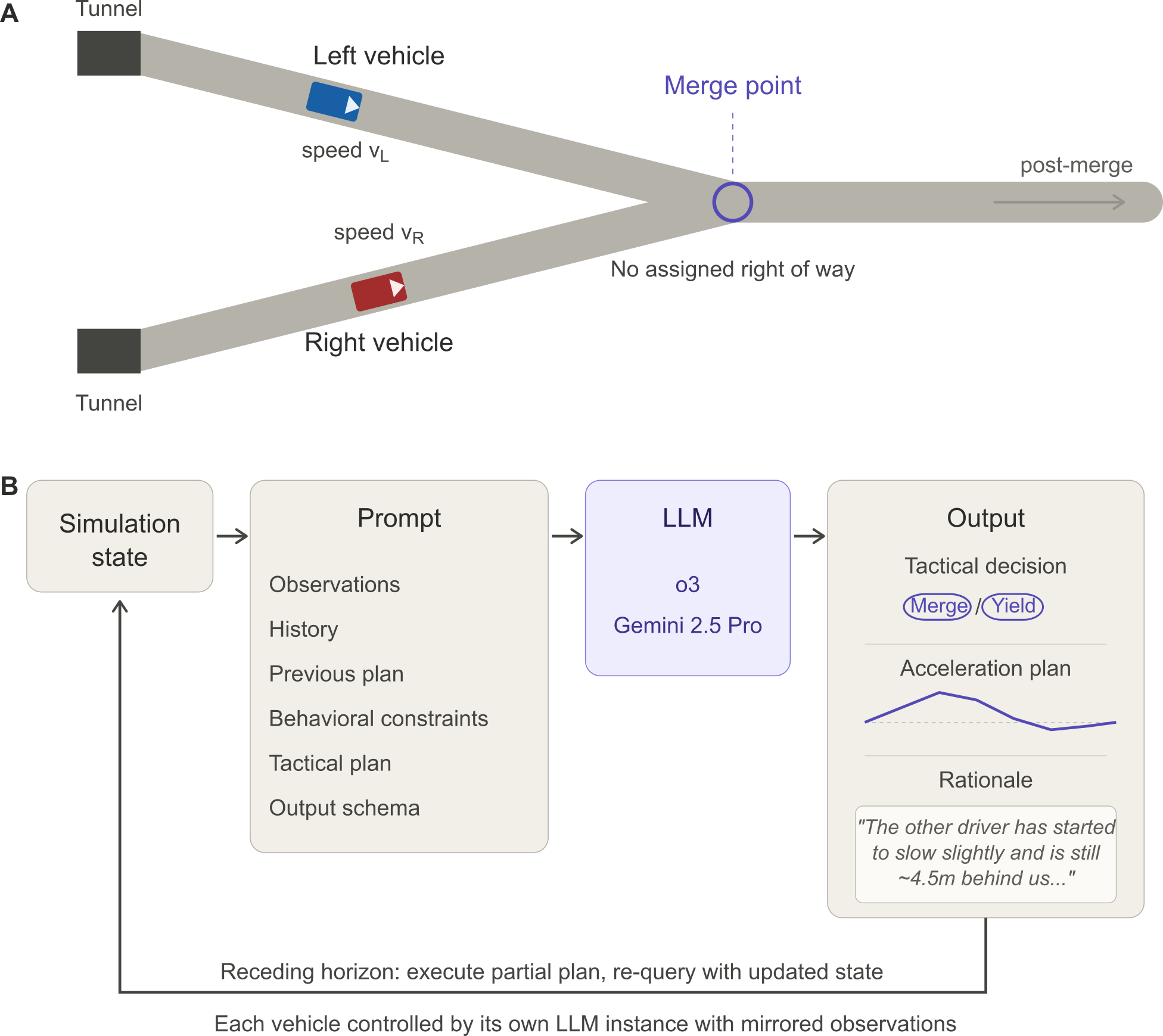}
    \caption{(A) The one-dimensional (1-D) merging task in the same simulated environment from~\cite{siebinga2024model}. Two vehicles emerge from tunnels on the branches of an inverse-Y track and approach a common merge point with no assigned right of way. Initial speeds ($v_{L}$, $v_{R}$) and projected headway at the merge are varied across 11 kinematic conditions, matching the human dataset of~\cite{siebinga2024human}. (B) Closed-loop LLM driver-agent framework. At each prompt time, the simulation state (panel A) is formatted into a structured prompt (components C0–C5, Table~\ref{tab:prompt_components_ablations}) and passed to a general-purpose LLM, which returns a tactical decision, an acceleration plan, and a natural-language rationale. The plan is executed under a receding horizon, after which the system re-queries the LLM with updated state. Both vehicles are controlled in parallel by independent LLM instances with mirrored observations.}
    \label{fig:simulation-framework}
\end{figure*}

Despite the rapid growth of research on LLM-based models of human driving behavior, from a scientific perspective such models remain insufficiently understood. The current literature is limited in at least two important respects, both of which limit the applicability of LLM-based models in AV safety assessment.

\textit{First}, much of the work emphasizes model development and benchmark performance, while evaluation of behavioral realism remains superficial. In the current literature, evaluation is typically based on high-level quantitative metrics, such as average displacement errors or collision rates. Although such metrics are informative, they provide limited insight into qualitative similarities and differences between model-generated and human driving behavior which is critical for meaningful AV safety assessment. 

\textit{Second}, research efforts usually focus on designing increasingly sophisticated model pipelines, yet it often remains unclear which components contribute to observed improvements. In particular, the capabilities of the core building blocks of such models -- pre-trained, general-purpose LLMs -- are rarely examined in isolation. Recent studies have begun to address this question in the context of simple car-following behavior~\cite{chen_genfollower_2024, gao_prompt-guided_2026}, but even then the evaluation against human behavior remains superficial. More interactive scenarios, such as merging or negotiation at intersections, remain largely unexplored in terms of systematically assessing what general-purpose LLMs can and cannot reproduce.

These two limitations are particularly important given known issues of LLM-based models which often generate plausible-looking outputs that have no relation to truth~\cite{hicks_chatgpt_2024}. Moreover, despite their ability to produce seemingly plausible explanations, LLMs remain fundamentally opaque black-box systems, limiting their applicability in safety-critical contexts. A deeper scientific understanding of generative AI-based driving behavior models is therefore essential for realizing their potential in AV safety assessment.

In this study, we aim to contribute to addressing these two limitations by: a) investigating general-purpose LLMs as minimal, standalone models of human driving behavior and b) rigorously evaluating them against human data. Rather than designing a complex architecture, we combine an engineered prompt with a publicly available pre-trained LLM and evaluate its performance in a simple interactive merging scenario (Figure~\ref{fig:simulation-framework}). Importantly, our objective is \textit{not to propose an LLM-based controller for AVs}, but to examine whether general-purpose LLMs can serve as \textit{models of human drivers} (as mentioned above, such models have important applications in AV safety assessment). By combining quantitative and qualitative analyses of two widely used LLMs against human data, we provide detailed insight into the strengths and limitations of general-purpose LLMs as standalone models of human driver behavior, and identify future improvements required for their viable use in AV safety assessment.
	
\section{Methods}
We implemented two LLM-controlled driving agents in the interactive one-dimensional (1-D) merging task (Figure~\ref{fig:simulation-framework}), matching the scenario and kinematic conditions previously used to obtain the human driving dataset~\cite{siebinga2024human}. We did not perform any task-specific training or parameter fitting, comparing the interaction outcomes under a fixed simulation and analysis pipeline. The objective is analogous to canonical car-following experiments in human factors research, where simplified environments are intentionally used to isolate behavioral dependencies before scaling to realistic traffic~\cite{saifuzzaman2014incorporating}. We evaluated two general-purpose LLMs with chain-of-thought reasoning enabled (OpenAI \textit{o3} and Google \textit{Gemini 2.5 Pro}) using (i) an engineered baseline prompt and (ii) seven single-factor prompt ablations. 

\subsection{Human dataset and driving task}
We used the dataset previously collected in a coupled driving simulator with a top-down view (see~\cite{siebinga2024human} for a detailed description of the dataset and the task). In each trial of the original experiment, two human participants interacted with each other on an inverse Y-shaped track (Figure~\ref{fig:simulation-framework}). The drivers independently but simultaneously controlled their own vehicle's acceleration with pedals while approaching a merge point. As there was no assigned right-of-way, participants had to resolve the merging conflict by dynamically negotiating via vehicle motion who goes first and who yields. Initial speeds of the two vehicles and the projected headway at the merge point were varied across conditions, producing 11 unique kinematic conditions. Vehicle headings were fixed (i.e., no steering was involved), and the experiment prevented explicit communication. Vehicles were modeled as point masses, using physical dimensions only for collision detection. 

\subsection{Simulation environment}
To simulate the LLM behavior in the simplified merging task, we reproduced the 11 kinematic conditions used in the human data. Two interacting agents were simulated in parallel using two instances of the LLM. Each agent selected a longitudinal acceleration $a_t$ bounded to $[-2.5,\,2.5]$~m/s$^2$. The simulator updated speed and position with a fixed step size $\Delta t = 0.2$~s (Table~\ref{tab:sim}). A collision was recorded when the center-to-center distance fell below the vehicle length. Each trial terminated when the first vehicle passed the end of the track or when a collision occurred.

\begin{table}[h]
\centering
\caption{Model and simulation parameters.}
\label{tab:sim}
\begin{tabular}{l c l c}
\hline
Plan horizon $T$ & $4.0$~s &
Memory window $T_m$ & $2.0$~s \\
Prompt rate $f_p$ & $1$~Hz &
Temperature $\gamma$ & $1.0$ [-] \\
Step $\Delta t$ & $0.2$~s & Vehicle length $l_{veh}$ & 4.5 m \\

\hline

\end{tabular}
\end{table}

\subsection{LLM agent interface and closed-loop control}
Each vehicle was controlled by an LLM policy (Figure~\ref{fig:simulation-framework}), queried at $f_p=1$~Hz (the value chosen to balance temporal resolution against computational cost and the associated environmental impact of LLM queries). At each prompt time $k$, an LLM agent received four inputs:
(1) \textbf{Current kinematic state}: ego speed, other speed, relative longitudinal distance, and distance-to-merge (when applicable);
(2) \textbf{Short-term history} over $T_m$: the ego’s executed accelerations and the other vehicle’s observed velocities;
(3) \textbf{Previous plan context}: the ego’s acceleration plan returned at the previous prompt;
(4) \textbf{Task context and constraints}: environment details, vehicle length, and behavioral guidelines (Sec.~\ref{sec:prompt}).

To ensure compatibility with the simulation framework, we required the LLM to output a fixed-length normalized acceleration plan $\hat{\mathbf{u}}\in[-1,1]^N$ over horizon $T$ (with $N=T/\Delta t$), which was subsequently mapped to physical acceleration via $a_t = a_{\max}\hat{u}_t$ (with $a_{\max}=2.5$~m/s$^2$), consistent with the simulator’s bounded acceleration control. The simulator then executed each agent’s plan in a receding-horizon loop. After each query, agents applied the first $1$~s segment of their plan (i.e., the first $1/\Delta t=5$ actions), and then re-queried the LLM with updated observations and the remaining (unexecuted) plan as context. Within each trial, both interacting vehicles were controlled independently by separate instances of the same LLM using the same prompt variant, with mirrored observations. We fixed the sampling temperature to $\gamma=1.0$ for all experiments to keep stochasticity constant across models, conditions, and prompt variants. Because temperature can change behavioral variability and tail-risk (including collision rates), our findings should be interpreted as conditional on $\gamma=1.0$; a systematic sweep is left for future work.

\subsection{Prompt template, ablations and sensitivity} \label{sec:prompt}
We engineered the baseline prompt variant following prior work using LLMs as (non-human) driving agents~\cite{cui2024receive} by combining (i) structured observations, (ii) explicit behavioral constraints, (iii) short-term memory to promote temporal consistency, and (iv) a strict, standardized output schema. For brevity, we summarized the template in Table~\ref{tab:prompt_components_ablations}; the full prompt text is provided in the \href{https://osf.io/pt4cd/overview?view_only=7902fdfba44c4ab48e1b75f1caae5080}{supplementary online material}.

To ensure reliable integration with the simulation framework, each response of the LLM was validated for schema compliance: the output had to contain a length-$N$ array with entries in $[-1,1]$. If validation failed, the agent used the most recent valid acceleration plan for the current planning interval.

We treat prompts as experimentally manipulable inductive biases; this enables controlled probing of behavioral structure of LLMs. Specifically, to identify prompt components essential for model behavior, we conducted a single-factor ablation study with seven prompt variants (Table~\ref{tab:prompt_components_ablations}), each removing exactly one prompt element or sub-element while keeping all remaining prompt content intact.

\begin{table}[h]
\centering
\caption{Prompt components (C) and single-factor ablations (A).}
\label{tab:prompt_components_ablations}
\setlength{\tabcolsep}{4pt}
\renewcommand{\arraystretch}{1.15}
\begin{tabular}{p{0.06\columnwidth} p{0.87\columnwidth}}
\toprule
\textbf{ID} & \textbf{Description (baseline)} \\
\midrule
C0 & \textbf{Observations and task context} (required): road type (merge vs.\ straight), vehicle length, relative distance/headway, ego and other velocities, and distance-to-merge when applicable. \\
C1 & \textbf{History} ($T_m$): past ego accelerations and past observed other-vehicle velocities. \\
C2 & \textbf{Previous acceleration plan}: acceleration plan returned at the previous prompt, included to encourage plan continuity under re-planning. \\
C3 & \textbf{Behavioral constraints}: collision avoidance, human-likeness, conservative distance keeping, and periodic reassessment. \\
C4 & \textbf{Tactical plan}: explicit merge-first vs.\ yield choice before generating the continuous control plan (merging scenarios only). \\
C5 & \textbf{Output schema} (required): optional M/Y token, 2--3 sentence rationale, and a length-$N$ normalized acceleration array $\hat{\mathbf{u}}\in[-1,1]^N$ returned in a Python code block (array only). \\
\midrule
\textbf{ID} & \textbf{Ablation}\\
\midrule
A1 & \textbf{(-C1)} Remove history. \\
A2 & \textbf{(-C2)} Remove previous acceleration plan. \\
A3 & \textbf{(-Safety)} Remove collision-avoidance instruction from C3. \\
A4 & \textbf{(-Human)} Remove human-likeness instruction from C3. \\
A5 & \textbf{(-Distance)} Remove distance/risk-aversion instruction from C3. \\
A6 & \textbf{(-Reassess)} Remove reassessment instruction from C3. \\
A7 & \textbf{(-C4)} Remove tactical plan. \\
\bottomrule
\end{tabular}
\end{table}

\subsection{Experimental design and analyzed samples}
We evaluated two general-purpose LLM driver agents, OpenAI \textit{o3} (o3-2025-04-16) and \textit{Gemini 2.5 Pro} (June 2025 release). For each model, the baseline prompt was simulated across all 11 kinematic conditions with 10 repetitions per condition (110 trials/model). To reduce API costs, each single-factor ablation was simulated with only 5 repetitions per condition (55 trials/ablation setting). For statistical comparisons of tactical decisions and safety margins, we excluded collisions and non-finished trials. This yielded $n=962$ human trials, $n=109$ \textit{o3} trials, and $n=82$ \textit{Gemini 2.5 Pro} trials for the analyses reported in Section~\ref{sec:baseline-results}.

\subsection{Behavioral metrics}
The following indicators were assessed for both human and simulated trials: \textbf{(1) Merge-first decision}: a binary variable indicating which vehicle enters the merged lane first (coded consistently with the experimental convention; collisions excluded).
\textbf{(2) Gap at merge}: the absolute inter-vehicle distance (meters) at the time the first vehicle reaches the merge point (collisions excluded).
\textbf{(3) Collision rate}: percentage of trials resulting in collision.
\textbf{(4) Speed deviation}: for each driver, the RMSE of velocity deviation from its initial velocity over the interaction window; we report the across-driver mean.
\textbf{(5) Joint contribution rate}: percentage of trials where both drivers’ deviation from the initial speed (measured by RMSE) is at least 0.5~m/s, indicating that both drivers substantially changed speed to resolve the merging conflict.

\subsection{Statistical analysis}
To quantify whether simulated behavior followed the same qualitative dependencies as humans, regression models were fit in Python using \texttt{statsmodels}. For merge-first probability we fit a logistic regression:
\begin{equation}
\mathrm{logit}\,P(\text{left merges first}) = \beta_0 + \beta_1 h + \beta_2 \Delta v,
\end{equation}
where $h$ is the signed projected headway at the merge point and $\Delta v$ is the signed initial relative velocity (definitions and signs are fixed across human and simulation analyses). For gap at merge $g$ we fit a linear regression using the same predictors as~\cite{siebinga2024human}:
\begin{equation}
g = \alpha_0 + \alpha_1 |h| + \alpha_2 |\Delta v| + \alpha_3(h \cdot \Delta v).
\end{equation}
Collisions were excluded from both regressions. For all analyses we used a significance level of $\alpha=0.05$.
	
\section{Results}
We report baseline-prompt behavior first, followed by the prompt ablation study. Figure~\ref{fig:baseline_summary} provides an overview of baseline operational control, tactical decisions, and safety margins. Figures and detailed tables for the ablation analysis are available in the \href{https://osf.io/pt4cd/overview?view_only=7902fdfba44c4ab48e1b75f1caae5080}{supplementary online material}.

\subsection{Baseline prompt performance} \label{sec:baseline-results}

\begin{figure*} [h]
    \centering
    \includegraphics[width=\linewidth]{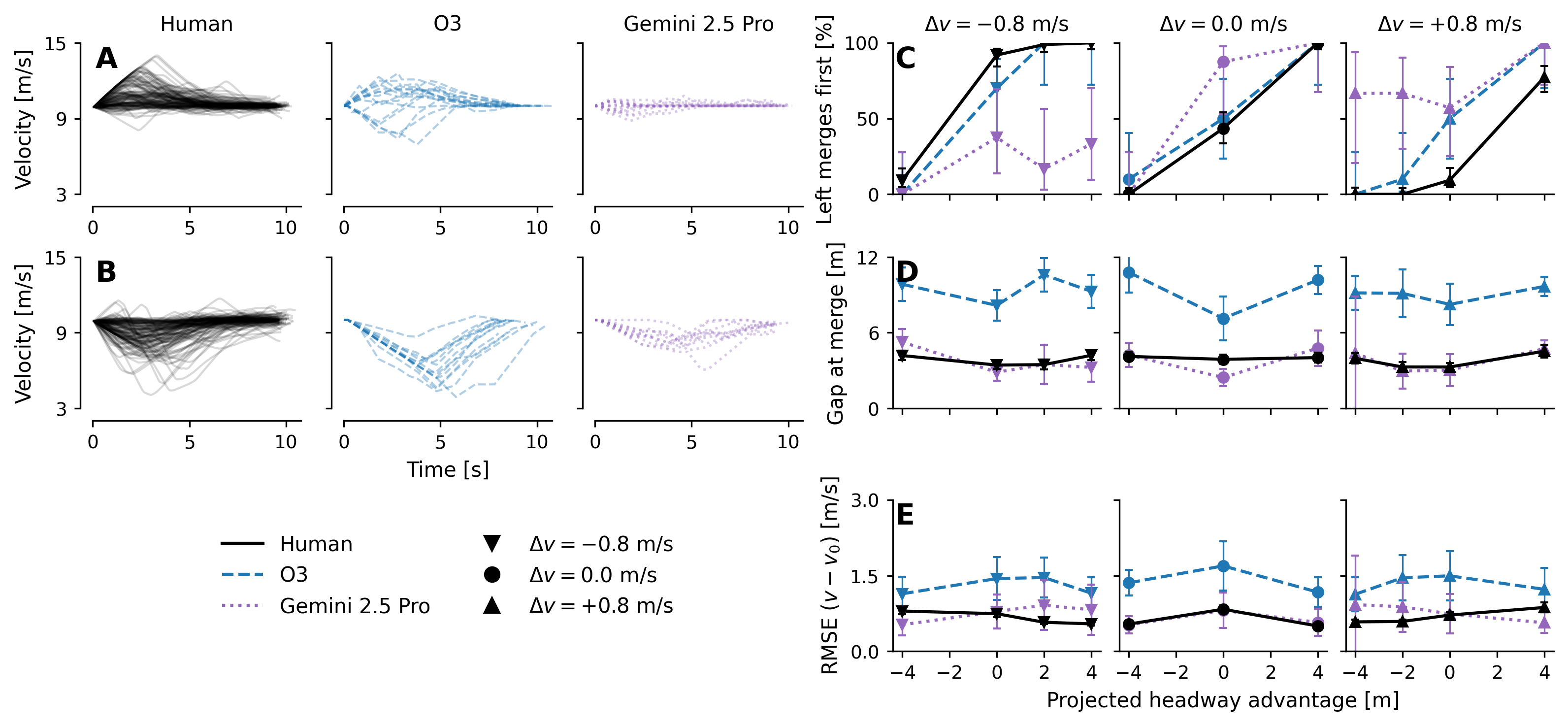}
    \caption{Baseline behavioral comparison between humans and LLM agents (collisions excluded).
   \textbf{ (A--B)} Left-vehicle velocity time series after exiting the tunnel ($t=0$), pooled across kinematic conditions, separated by whether the left vehicle merges first \textbf{(A)} or second \textbf{(B)}, illustrating piecewise-linear velocity profiles characteristic of intermittent acceleration control.
    \textbf{(C)} Probability that the left vehicle merges first as a function of projected headway advantage, shown separately for each initial relative-velocity condition $\Delta v \in \{-0.8, 0.0, +0.8\}$ m/s.
    \textbf{(D)} Mean gap at the merge point versus projected headway advantage, per $\Delta v$ condition.
    \textbf{(E)} Mean per-vehicle RMSE of velocity deviation from the initial velocity versus projected headway advantage, per $\Delta v$ condition (each trial contributes two RMSE values: left and right vehicle).
    Error bars indicate 95\% confidence intervals (\textbf{C}: Wilson intervals for proportions; \textbf{D--E}: t-based intervals for means).}
    \label{fig:baseline_summary}
\end{figure*}

\textbf{Tactical decisions (who merges first).}
In the human data, the probability of a vehicle merging first increases with its projected headway advantage and decreases with relative velocity advantage (provided that the projected headway advantage remains the same). Both LLM agents reproduce the qualitative headway dependency (Table~\ref{tab:who_first_comparison}): projected headway is a strong positive predictor for humans ($b=1.11$, $p<.001$), \textit{o3} ($b=1.11$, $p<.001$), and \textit{Gemini 2.5 Pro} ($b=0.50$, $p<.001$), although the latter model does deviate substantially from the human data at large headway advantage levels (Figure~\ref{fig:baseline_summary}C). However, the relative-velocity effect diverged from human data for both models. Human drivers are less likely to merge first with increasing relative velocity advantage ($b=-3.30$, $p<.001$). In contrast, \textit{o3} shows no evidence of a significant relative-velocity effect ($b=-0.57$, $p=.253$) while \textit{Gemini 2.5 Pro} exhibits an opposite effect ($b=2.12$, $p<.001$).

\textbf{Safety margins (gap at merge).}
In the human data, the gap at the merge point increases with the magnitude of projected headway and is not explained by the magnitude of initial relative velocity. This qualitative pattern also holds for both LLM agents (Table~\ref{tab:gap_at_merge_comparison}; Figure~\ref{fig:baseline_summary}D): the coefficient on absolute projected headway is positive and significant for humans, \textit{o3}, and \textit{Gemini 2.5 Pro}, whereas absolute relative velocity is non-significant across all three. The agents differ in safety margin magnitude at moment of merging. \textit{o3} maintains substantially larger gaps on average (9.28~m) than humans (3.85~m) and \textit{Gemini 2.5 Pro} (3.84~m), indicating more conservative behavior (Table~\ref{tab:behavior_indicators_by_source}).

\textbf{Operational control and high-level metrics.}
Similar to humans, both models tend to produce piecewise-linear velocity profiles (Figure~\ref{fig:baseline_summary}A and B) which is characteristic of intermittent acceleration control~\cite{siebinga2024human}. Operational and joint-interaction indicators are presented in Table~\ref{tab:behavior_indicators_by_source}: \textit{o3} produces no collisions but operates with larger deviations from initial velocity (RMSE 1.34~m/s) and frequent joint contribution to resolve the merging conflict (94.5\%; inconsistent with human data). \textit{Gemini 2.5 Pro} matches the human average gap closely, is relatively consistent with human joint contribution rate, shows similar speed deviations (0.70~m/s vs.\ 0.66~m/s) but collides frequently (25.45\%). 

\subsection{Prompt ablations and cross-model sensitivity}
Table~\ref{tab:ablation_qual_combined} reports alignment of model behavior with qualitative indicators of human behavior across single-factor removals, and Table~\ref{tab:ablation_quant_combined} reports quantitative metrics. Across ablations, \textit{o3} maintains comparatively stable qualitative alignment (typically 4/5), whereas \textit{Gemini 2.5 Pro} degrades more strongly under several removals, particularly those affecting memory structure.

\textbf{Qualitative sensitivity.}
For \textit{o3}, A6 (\texttt{-Reassess}) reduces tactical behavior alignment: merge-first probability no longer depends on projected headway. In contrast, A2 (\texttt{-Prev.\ acc.\ plan}) and A7 (\texttt{-Tactical plan}) improve \textit{o3}'s tactical alignment by restoring the human-like dependency on relative velocity advantage (5/5). \textit{Gemini 2.5 Pro} is more sensitive to memory-related structure. A2 (\texttt{-Prev.\ acc.\ plan}) disrupts piecewise-linear velocities and weakens gap-related indicators, while A1 (\texttt{-History}) also degrades gap-related alignment.

\textbf{Quantitative tradeoffs.}
Across ablations, \textit{o3}'s collision rate increases from 0\% toward the human rate (2.83\%), with the largest increases under A1 (\texttt{-History}) and A2 (\texttt{-Prev.\ acc.\ plan}). For \textit{Gemini 2.5 Pro}, the baseline collision rate is high (25.45\%); A2 (\texttt{-Prev.\ acc.\ plan}) and A4 (\texttt{-Human}) reduce collisions to 16.36\% and 14.55\%, respectively, but with tradeoffs in other indicators (e.g., increased gap deviations under A2). Across both models, average speed deviation from initial velocity is relatively insensitive to ablations, whereas joint contribution rate changes substantially under specific removals (e.g., \textit{Gemini 2.5 Pro} under A1 \texttt{-History}). Across both models, no single ablation uniformly improves alignment with human behavior; instead, each removal reveals model-specific tradeoffs between qualitative alignment and safety performance.

\begin{table} [h]
\centering
\caption{Logistic regression models of which driver merged first. Human (H): n=962; o3: n=109; Gemini 2.5 Pro (G): n=82.}
\label{tab:who_first_comparison}
\begin{tabular}{lrrrr}
\hline
\textbf{} & \textbf{b} & \textbf{SE} & \textbf{Z} & \textbf{p} \\
\hline
Intercept - H & -0.30 & 0.135 & -2.26 & 0.024 \\
Intercept - o3 & 0.44 & 0.340 & 1.29 & 0.198 \\
Intercept - G & 0.08 & 0.301 & 0.26 & 0.792 \\
Projected headway - H & 1.11 & 0.075 & 14.89 & $< .001$ \\
Projected headway - o3 & 1.11 & 0.243 & 4.56 & $< .001$ \\
Projected headway - G & 0.50 & 0.125 & 3.97 & $< .001$ \\
Relative velocity - H & -3.30 & 0.310 & -10.64 & $< .001$ \\
Relative velocity - o3 & -0.57 & 0.503 & -1.14 & 0.253 \\
Relative velocity - G & 2.12 & 0.534 & 3.97 & $< .001$ \\
\hline
\end{tabular}
\end{table}

\begin{table} [!htbp]
\centering
\caption{Linear regression models of the gap that drivers keep between the vehicles at the merge point. Human (H): n=962; o3: n=109; Gemini 2.5 Pro (G): n=82.}
\label{tab:gap_at_merge_comparison}
\begin{tabular}{lrrrr}
\hline
\textbf{} & \textbf{Est.} & \textbf{SE} & \textbf{t} & \textbf{p} \\
\hline
Intercept - H & 3.54 & 0.146 & 24.30 & $< .001$ \\
Intercept - o3 & 8.14 & 0.483 & 16.86 & $< .001$ \\
Intercept - G & 2.73 & 0.335 & 8.16 & $< .001$ \\
Abs. projected headway - H & 0.17 & 0.034 & 5.00 & $< .001$ \\
Abs. projected headway - o3 & 0.46 & 0.114 & 4.03 & $< .001$ \\
Abs. projected headway - G & 0.40 & 0.082 & 4.95 & $< .001$ \\
Abs. relative velocity - H & -0.20 & 0.166 & -1.19 & 0.236 \\
Abs. relative velocity - o3 & -0.03 & 0.554 & -0.05 & 0.959 \\
Abs. relative velocity - G & 0.05 & 0.376 & 0.13 & 0.894 \\
Headway : relative velocity - H & 0.06 & 0.029 & 2.16 & 0.031 \\
Headway : relative velocity - o3 & 0.02 & 0.098 & 0.21 & 0.834 \\
Headway : relative velocity - G & 0.21 & 0.072 & 2.89 & 0.005 \\
\hline
\end{tabular}
\end{table}

\begin{table*}[h]
\centering
\caption{Selected qualitative and quantitative indicators of human (H) behavior during merging conflicts.}
\label{tab:behavior_indicators_by_source}
\begin{tabular}{lccc}
\hline
\textbf{Indicator} & \textbf{H} & \textbf{o3} & \textbf{Gemini 2.5 Pro} \\
\hline

\multicolumn{4}{l}{\textbf{Qualitative}} \\

The probability of merging first increases with projected headway advantage. & $\checkmark$ & $\checkmark$ & $\checkmark$ \\
The probability of merging first decreases with relative velocity advantage. & $\checkmark$ & \textbf{X} & \textbf{X} \\
The safety gap size is moderately positively affected by projected headway. & $\checkmark$ & $\checkmark$ & $\checkmark$ \\
The safety gap size is not affected by initial relative velocity difference. & $\checkmark$ & $\checkmark$ & $\checkmark$ \\
Drivers resolve the merging conflict with intermittent, piecewise-linear velocity profiles. & $\checkmark$ & $\checkmark$ & $\checkmark$ \\
\textbf{Total qualitative score.} & \textbf{5/5} & \textbf{4/5} & \textbf{4/5 }\\ \hline
\multicolumn{4}{l}{\textbf{Quantitative}} \\

Collision rate (\%). & 2.83 & 0.00 & 25.45 \\
Average gap at merge (m). & 3.85 & 9.28 & 3.84 \\
Average RMSE of velocity deviation from initial velocity (m/s). & 0.66 & 1.34 & 0.70 \\
The rate of drivers jointly contributing to safety gap (each RMSE $\geq$ 0.5 m/s from initial velocity) (\%). & 53.0 & 94.5 & 42.7 \\

\hline
\end{tabular}
\end{table*}

\begin{table*}[h]
\centering
\caption{Alignment of model behavior with qualitative indicators (I) of human behavior across seven prompt ablations. \\
I1: headway$\uparrow$ merge-first; I2: $\Delta v\uparrow$ merge-first$\downarrow$; I3: headway$\uparrow$ gap; \\ I4: gap $\perp\,\Delta v$; I5: piecewise-constant control. S: total score (out of 5).}
\label{tab:ablation_qual_combined}
\setlength{\tabcolsep}{4pt}
\renewcommand{\arraystretch}{1.2}
\begin{tabular}{lcccccc cccccc}
\toprule
& \multicolumn{6}{c}{\textbf{o3}} & \multicolumn{6}{c}{\textbf{Gemini 2.5 Pro}} \\
\cmidrule(lr){2-7}\cmidrule(lr){8-13}
\textbf{Setting} & \textbf{I1} & \textbf{I2} & \textbf{I3} & \textbf{I4} & \textbf{I5} & \textbf{S}
                & \textbf{I1} & \textbf{I2} & \textbf{I3} & \textbf{I4} & \textbf{I5} & \textbf{S} \\
\midrule
Baseline                & \checkmark & \textbf{X} & \checkmark & \checkmark & \checkmark & \textbf{4/5}
                        & \checkmark & \textbf{X} & \checkmark & \checkmark & \checkmark & \textbf{4/5} \\
A1 (-History)           & \checkmark & \textbf{X} & \checkmark & \checkmark & \checkmark & \textbf{4/5}
                        & \checkmark & \textbf{X} & \textbf{X} & \checkmark & \checkmark & \textbf{3/5} \\
A2 (-Prev.\ acc.\ plan) & \checkmark & \checkmark & \checkmark & \checkmark & \checkmark & \textbf{5/5}
                        & \checkmark & \textbf{X} & \textbf{X} & \checkmark & \textbf{X} & \textbf{2/5} \\
A3 (-Safety)            & \checkmark & \textbf{X} & \checkmark & \checkmark & \checkmark & \textbf{4/5}
                        & \checkmark & \textbf{X} & \checkmark & \checkmark & \checkmark & \textbf{4/5} \\
A4 (-Human)             & \checkmark & \textbf{X} & \checkmark & \checkmark & \checkmark & \textbf{4/5}
                        & \checkmark & \textbf{X} & \checkmark & \checkmark & \checkmark & \textbf{4/5} \\
A5 (-Distance)          & \checkmark & \checkmark & \checkmark & \textbf{X} & \checkmark & \textbf{4/5}
                        & \checkmark & \textbf{X} & \checkmark & \checkmark & \checkmark & \textbf{4/5} \\
A6 (-Reassess)          & \textbf{X} & \textbf{X} & \checkmark & \checkmark & \checkmark & \textbf{3/5}
                        & \checkmark & \textbf{X} & \checkmark & \checkmark & \checkmark & \textbf{4/5} \\
A7 (-Tactical plan)     & \checkmark & \checkmark & \checkmark & \checkmark & \checkmark & \textbf{5/5}
                        & \checkmark & \textbf{X} & \textbf{X} & \checkmark & \checkmark & \textbf{3/5} \\
\bottomrule
\end{tabular}
\end{table*}

\begin{table*}[h]
\centering
\caption{Quantitative performance indicators of the models across prompt ablations. \\
Human reference values: collision=2.83\%, gap=3.85~m, RMSE=0.66~m/s, joint rate=53.0\%.}
\label{tab:ablation_quant_combined}
\setlength{\tabcolsep}{4pt}
\renewcommand{\arraystretch}{1.05}
\begin{tabular}{lcccc cccc}
\toprule
& \multicolumn{4}{c}{\textbf{o3}} & \multicolumn{4}{c}{\textbf{Gemini 2.5 Pro}} \\
\cmidrule(lr){2-5}\cmidrule(lr){6-9}
\textbf{Setting} &
\textbf{Coll. (\%)} & \textbf{Gap (m)} & \textbf{RMSE} & \textbf{Joint (\%)} &
\textbf{Coll. (\%)} & \textbf{Gap (m)} & \textbf{RMSE} & \textbf{Joint (\%)} \\
\midrule
Baseline                & 0.00 & 9.28 & 1.34 & 94.50 & 25.45 & 3.84 & 0.70 & 42.68 \\
A1 (-History)           & 2.73 & 9.29 & 1.30 & 95.33 & 38.18 & 4.82 & 0.85 & 76.47 \\
A2 (-Prev.\ acc.\ plan) & 2.73 & 10.63 & 1.50 & 100.00 & 16.36 & 5.14 & 0.86 & 47.83 \\
A3 (-Safety)            & 3.64 & 9.06 & 1.24 & 98.11 & 30.91 & 3.68 & 0.63 & 47.37 \\
A4 (-Human)             & 1.82 & 9.47 & 1.32 & 98.15 & 14.55 & 4.39 & 0.76 & 41.49 \\
A5 (-Distance)          & 0.91 & 9.42 & 1.25 & 98.17 & 26.36 & 3.63 & 0.63 & 40.74 \\
A6 (-Reassess)          & 0.91 & 9.62 & 1.28 & 96.33 & 27.27 & 3.58 & 0.68 & 52.50 \\
A7 (-Tactical plan)     & 0.91 & 9.57 & 1.33 & 99.08 & 38.18 & 3.72 & 0.61 & 52.94 \\
\bottomrule
\end{tabular}
\end{table*}
	
\section{Discussion}
This study set out to examine what general-purpose LLMs can and cannot capture as standalone models of human driving behavior for the simple case of a simulated interactive merging task. Given the rapid integration of LLMs into driving behavior modeling pipelines for AV safety assessment, understanding their intrinsic properties is scientifically necessary. As a modeling paradigm, general-purpose LLMs occupy a unique position: they offer unprecedented flexibility without the need for scenario-specific calibration or domain-specific training data. Whether this flexibility translates into behaviorally valid driver models is the question this study poses.

Several findings indicate that general-purpose LLMs can reproduce meaningful aspects of human interactive driving in our considered task. Both models (OpenAI \textit{o3} and Google \textit{Gemini 2.5 Pro}) replicate the human dependency of merge-first decisions on projected headway, and both generate piecewise-linear velocity profiles consistent with human operational behavior. For \textit{o3}, the best-performing prompt configurations align with five out of five qualitative indicators simultaneously in simulations spanning all 11 kinematic conditions. Remarkably, these results were obtained in a zero-shot setting, without any training on the modeled data.

However, the same experiments expose systematic boundaries. Neither model consistently reproduces the human-like effect of relative velocity on merge order. One possible explanation for this failure is that text-based state representations are insufficient for supporting this type of reasoning over both projected headway and relative velocity advantages. This would be consistent with limitations of current LLMs arising from interaction between textual state representations and autoregressive reasoning~\cite{quattrociocchi2025epistemological}. Alternative explanations (e.g., biases in the LLMs' training data) remain plausible and require future study. 

Safety performance of the models also diverges sharply: \textit{o3} resolves conflicts too conservatively (large safety margins; no collisions), while \textit{Gemini 2.5 Pro} matches human average gaps but collides frequently, demonstrating that on average human-like behavior can co-occur with unreliable interaction safety. Our ablation study further revealed that prompt components function as model-specific inductive biases. For \textit{o3}, removing plan-anchoring elements enables (human-like) sensitivity to relative velocity, suggesting that plan continuity can overweigh static cues at the expense of dynamic ones. For \textit{Gemini 2.5 Pro}, memory-related structure is critical for maintaining control coherence. These asymmetric effects mean that prompt design choices do not transfer across model families, arguing for structured sensitivity testing whenever behavioral claims about LLMs are made.

These findings have potential implications for the use of LLMs as human behavior models in AV safety assessment. The ability to reproduce qualitative behavioral dependencies without finetuning is promising for traffic simulation, but the failure to consistently capture velocity advantage dependencies and the strong model-prompt coupling indicate that current general-purpose LLMs cannot yet be treated as validated human behavior models. However, much further research in this direction is needed to further understand LLM capabilities as standalone human driver behavior models. For instance, the key limitation of this study is the simplicity of the 1-D scenario. While the underlying human data~\cite{siebinga2024human} has been successfully used in previous work to develop mechanistic models~\cite{siebinga2024model} which subsequently generalized well to realistic scenarios~\cite{siebinga_modeling_2025}, its simplicity necessitates extending our evaluation to scenarios with additional degrees of freedom (2-D merges, lane changes), even if the latter would likely require more sophisticated prompting approaches. Addressing this as well as other limitations (perfect observations, the modest number of LLM repetitions relative to the human dataset, and the use of a single temperature setting) will clarify whether the boundaries identified here are fundamental to current LLM architectures or addressable through richer state representations and prompting strategies.

In addition to practical implications for AV safety assessment, our findings add to the recent fundamental literature on LLMs that challenges the hegemony of mechanistic models as faithful representations of human cognition. In particular, recent work demonstrated that fine-tuned LLMs can predict and simulate fine-grained details of human cognition (such as cognitive biases and response times) across a large number of cognitive tasks~\cite{binz_foundation_2025}. While our study is narrower in scope, it uniquely demonstrates the potential of LLMs in capturing fine-grained details of human behavior in dynamic interactive tasks -- something that traditional cognitive models typically struggle with~\cite{carvalho_naturalistic_2025}.

\section{Conclusion}
General-purpose LLMs promise an appealing paradigm for human modeling: a single model deployable without scenario-specific calibration. We evaluated two LLMs (\textit{o3} and \textit{Gemini 2.5 Pro}) as standalone closed-loop agents in a simplified interactive merging task and compared their behavior against human data. Both reproduced key human-like spatial dependencies and coherent control patterns in a zero-shot setting. However, systematic limitations emerged in capturing the dependencies of behavior on initial kinematic conditions and in the strong sensitivity to prompt structure. These findings indicate that while general-purpose LLMs can capture meaningful aspects of human interactive driving in minimal settings, their validity as human behavior models remains conditional and requires further validation in richer scenarios.

\section*{ACKNOWLEDGMENTS}
We thank the Transport \& Mobility Institute of Delft University of Technology for funding this research.
	\bibliographystyle{IEEEtran}
	\bibliography{root} 
	
\end{document}